\documentclass[lettersize,journal]{IEEEtran}
\usepackage[table]{xcolor}
\usepackage{amsmath,amsfonts}
\usepackage{amssymb}
\usepackage{amsmath}
\usepackage{algorithmic}
\usepackage{algorithm}
\usepackage{array}

\usepackage{textcomp}
\usepackage{stfloats}
\usepackage{url}
\usepackage{verbatim}
\usepackage{graphicx,subcaption}
\usepackage{adjustbox}
\usepackage{cite}
\usepackage{mathtools}
\usepackage{multirow}
\usepackage{listings}
\usepackage{xcolor}
\usepackage{booktabs}

\graphicspath{{./Figures/}} 

\usepackage{newfloat}
\usepackage{listings}
\lstdefinestyle{mystyle}{
    backgroundcolor=\color{white},   
    commentstyle=\color{gray},
    keywordstyle=\color{blue},       
    numberstyle=\tiny\color{gray},
    stringstyle=\color{purple},
    basicstyle=\ttfamily\footnotesize,
    breakatwhitespace=false,         
    breaklines=true,                 
    captionpos=b,                    
    keepspaces=true,                 
    numbers=none,                    
    numbersep=5pt,                  
    showspaces=false,                
    showstringspaces=false,
    showtabs=false,                  
    tabsize=2
}
\DeclareCaptionStyle{ruled}{labelfont=normalfont,labelsep=colon,strut=off} 
\begin{document}

\title{Customize Segment Anything Model for Multi-Modal Semantic Segmentation \\ with Mixture of LoRA Experts}
\author{Chenyang Zhu, Bin Xiao, Lin Shi, Shoukun Xu, Xu Zheng$^{\dagger}$\thanks{$\dagger$ Corresponding Author}
\thanks{


  Xu Zheng is with the AI Thrust, HKUST(GZ), Guangdong, China (E-mail: zhengxu128@gmail.com).}
}

\markboth{Journal of \LaTeX\ Class Files,~Vol.~14, No.~8, August~2021}%
{Shell \MakeLowercase{\textit{et al.}}: A Sample Article Using IEEEtran.cls for IEEE Journals}

\maketitle

\begin{figure*}[htp]
  \begin{center}
     \includegraphics[width=\linewidth]{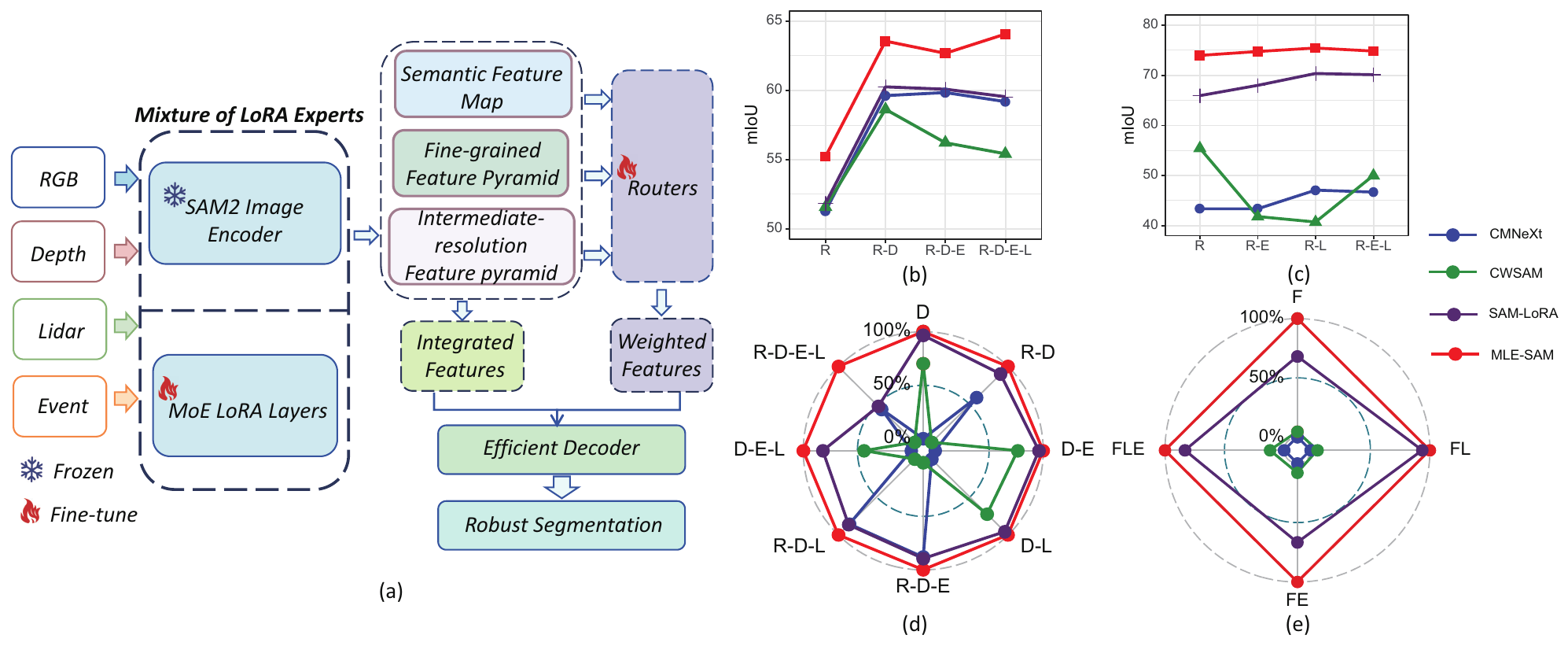}
     \caption{
     (a)Overall of MLE-SAM, (b) Performance on DELIVER (R-D-E-L Modalities), (c) Performance on MUSES (F-E-L Modalities), (d) Evaluation Across Modality Combinations and Scenarios on DELIVER, and (e) on MUSES Datasets.}\label{fig:teaser}
  \end{center}
\end{figure*}

\begin{abstract}  
The recent Segment Anything Model (SAM) represents a significant breakthrough in scaling segmentation models, delivering strong performance across various downstream applications in the RGB modality. However, directly applying SAM to emerging visual modalities, such as depth and event data results in suboptimal performance in multi-modal segmentation tasks. In this paper, we make the first attempt to adapt SAM for multi-modal semantic segmentation by proposing a Mixture of Low-Rank Adaptation Experts (MoE-LoRA) tailored for different input visual modalities. By training only the MoE-LoRA layers while keeping SAM’s weights frozen, SAM’s strong generalization and segmentation capabilities can be preserved for downstream tasks. Specifically, to address cross-modal inconsistencies, we propose a novel MoE routing strategy that adaptively generates weighted features across modalities, enhancing multi-modal feature integration. Additionally, we incorporate multi-scale feature extraction and fusion by adapting SAM’s segmentation head and introducing an auxiliary segmentation head to combine multi-scale features for improved segmentation performance effectively.
Extensive experiments were conducted on three multi-modal benchmarks: DELIVER, MUSES, and MCubeS. The results consistently demonstrate that the proposed method significantly outperforms state-of-the-art approaches across diverse scenarios. Notably, under the particularly challenging condition of missing modalities, our approach exhibits a substantial performance gain, achieving an improvement of 32.15\% compared to existing methods.
\end{abstract}


\section{Introduction}
Accurate segmentation of diverse objects is pivotal for various scene understanding applications, including robotic perception, autonomous driving, and AR/VR~\cite{shi2000normalized,minaee2021image}. The Segment Anything Model (SAM)\cite{kirillov2023segment} represents a groundbreaking advancement in instance segmentation, particularly for RGB images. Trained on an extensive dataset of 11 million high-resolution images and over 1 billion annotated segmentation masks, SAM achieves exceptional zero-shot segmentation performance, enabling its application across diverse domains such as medical imaging, remote sensing, and more\cite{gu2024build,wu2023medical,sun2023single,yan2023ringmo}.

While SAM has revolutionized single-modality segmentation tasks, particularly for RGB images, its application to multi-modal segmentation presents unique challenges. Emerging domains often require integrating diverse modalities such as depth and event data, which capture complementary scene information but exhibit distinct characteristics from RGB data. Furthermore, the recently proposed SAM2 model~\cite{ravi2024sam} incorporates temporal dimensions for video segmentation, addressing additional complexities such as motion, deformation, occlusion, and lighting variations. These advancements extend SAM's applicability to dynamic and multi-modal environments, but integrating cross-modal information while preserving SAM’s generalization capabilities remains under-explored.


Despite its success in single-modality segmentation, extending SAM to multi-modal semantic segmentation poses significant challenges. Each modality, such as LiDAR, radar, and event cameras, exhibits distinct spatial, temporal, and noise characteristics, complicating their seamless integration into SAM's architecture~\cite{luo2024zero}. SAM’s pre-trained features, optimized for RGB images, often result in suboptimal performance when directly applied to heterogeneous multi-modal data. Real-world scenarios further complicate this integration, as missing or unreliable modalities can degrade performance, and SAM lacks mechanisms to adaptively handle incomplete inputs~\cite{brodermann2024muses,zheng2024centering,zheng2025learning}. Additionally, effective multi-modal fusion requires advanced techniques to align, weigh, and integrate inputs while preserving the complementary strengths of each modality. Achieving robust fusion requires addressing several challenges, including mitigating modality-specific noise, harmonizing discrepancies in spatial and temporal resolutions, and balancing the contributions of each input modality~\cite{zhang2023delivering}. 

In this work, we present a novel framework that extends SAM2’s functionality to support multi-modal semantic segmentation. As shown in Figure~\ref{fig:teaser}(a), our approach incorporates Low-Rank Adaptation (LoRA) modules designed for each modality, facilitating efficient modality-specific fine-tuning while preserving the generalization capabilities of SAM2’s pre-trained image encoder. To address the inherent challenges of multi-modal fusion, we develop a Mixture of LoRA Experts (MLE) routing mechanism that adaptively generates weighted feature representations, ensuring effective integration across modalities and mitigating inconsistencies caused by noise or missing inputs. 
Meanwhile, we enhance the SAM2 segmentation pipeline by incorporating multi-scale feature extraction and fusion mechanisms. Specifically, we augment the original segmentation head with an auxiliary head designed to exploit complementary information across multiple scales, leading to improved segmentation accuracy. 

Extensive experiments conducted on benchmark datasets, including DELIVER~\cite{zhang2023delivering}, MUSES~\cite{brodermann2024muses}, and MCubeS~\cite{liang2022multimodal}, demonstrate the superior performance of our framework in multi-modal semantic segmentation tasks. As illustrated in Figure~\ref{fig:teaser}(b) and (c), our approach achieves a significant improvement of +4.9\% on the DELIVER dataset with four modalities and +28.14\% on the MUSES dataset with three modalities, compared to state-of-the-art methods. 
Detailed ablation studies confirm the individual contributions of each module to the overall performance. Furthermore, additional experiments under challenging conditions, such as noisy or missing modalities, highlight the robustness and adaptability of the proposed model, emphasizing its practical utility in real-world scenarios. Notably, as shown in Figure~\ref{fig:teaser}(d) and (e), our model achieves a performance gain of \textbf{14.13\%} on the DELIVER dataset and \textbf{32.15\%} on the MUSES dataset in these adverse settings, further establishing its efficacy and reliability.

Our contributions are outlined as follows:
\textbf{(I)} We improve the SAM2 framework by integrating a MoE mechanism with LoRA modules for multi-modal semantic segmentation tasks. This design enables efficient modality-specific adaptation by training distinct LoRA modules for each modality and leveraging a dynamic routing mechanism to integrate features across modalities effectively.
\textbf{(II)} We redesign the SAM2 segmentation pipeline by incorporating a modified segmentation head tailored for multi-modal input and introducing an auxiliary segmentation head. This configuration facilitates the effective fusion of multi-scale features, significantly improving segmentation accuracy.
\textbf{(III)} Our method achieves state-of-the-art performance on three widely-used multi-modal benchmarks, ranging from synthetic to real-world scenarios, surpassing existing methods in terms of segmentation accuracy and generalization across diverse modalities.
\textbf{(IV)} Extensive experimental evaluation demonstrates the robustness of the proposed framework under challenging conditions, including missing modalities and high levels of noise. The results highlight its adaptability and reliability for real-world applications.
  
\section{Related Work}\label{sec:relatedwork}

\subsection{Multi-modal Semantic Segmentation}
Multi-modal semantic segmentation seeks to leverage complementary information from multiple sensing modalities, such as RGB, depth, and thermal data, to assign semantic labels to each pixel, thereby improving the accuracy and robustness of scene understanding~\cite{zhang2021deep}. This task is predominantly addressed using encoder-decoder architectures, where the encoder extracts hierarchical features, and the decoder reconstructs pixel-level predictions~\cite{ronneberger2015u,badrinarayanan2017segnet,wang2022unetformer}. 

The evolution of encoders has been significantly influenced by Fully Convolutional Networks (FCNs), which enable end-to-end learning for pixel-level predictions~\cite{long2015fully,tian2021class}. Notable advancements in FCNs include the introduction of dilated convolutions to expand the receptive field~\cite{wu2019fastfcn,gao2023rethinking} and pyramid pooling modules to incorporate multi-scale contextual information~\cite{liu2019feature}. DeepLab further refined these methods by combining atrous convolutions with fully connected conditional random fields to enhance segmentation boundaries and accuracy~\cite{chen2017deeplab}. However, FCNs face challenges in capturing long-range dependencies, which are essential for understanding complex scenes. Transformer-based encoders address this limitation by employing self-attention mechanisms to model global context effectively~\cite{zheng2021rethinking,strudel2021segmenter,xie2021segformer,nguyen2022evaluating,chen2021transunet,shi2023transformer,zhou2022survey}. Moreover, transformer-based decoders integrate robust multi-level context mining and process diverse multi-scale features extracted by the encoder, enabling precise and efficient segmentation, particularly in complex or high-resolution images~\cite{han2022survey,shi2022transformer,panboonyuen2021transformer,zheng2024learning}.

Combining information from different modalities enhances scene understanding in multi-modal segmentation, especially in challenging environments where a single modality may be insufficient. Early fusion strategies combine data from all modalities at the input level, allowing the encoder to learn joint representations but risking redundancy or noise in the fused input~\cite{hazirbas2017fusenet,ding2019acnet,sun2019rtfnet}. In contrast, late fusion methods process each modality independently before combining features during decoding. This preserves modality-specific characteristics but may limit inter-modal interactions~\cite{valada2017adapnet,cheng2017locality,diakogiannis2020resunet}. Adaptive fusion strategies, which dynamically integrate multi-modal data at various stages of the network, have emerged as a flexible solution. These approaches refine features across modalities at different abstraction levels, often incorporating cross-modal attention mechanisms or specialized modules to enhance feature interactions~\cite{zhou2022cimfnet,ma2023adjacent,he2023multimodal,ma2024multilevel}.

\subsection{SAM for Semantic Segmentation}
SAM~\cite{kirillov2023segment} and DINO v2~\cite{oquab2023dinov2} are prominent foundation models for image segmentation, leveraging Vision Transformers as their backbone. SAM includes a mask decoder and a flexible prompt encoder that supports diverse inputs, such as points, bounding boxes, and text, enabling zero-shot instance segmentation. Despite its versatility, SAM faces challenges in semantic segmentation due to its training on large-scale datasets focused on object boundaries rather than semantic labels~\cite{li2023semantic}. 
To adapt SAM for semantic segmentation, ClassWise-SAM-Adapter (CWSAM) introduces lightweight adapters, a classwise mask decoder, and efficient task-specific input preprocessing to assign semantic labels in challenging SAR imagery efficiently~\cite{pu2024classwise}. The SAM-to-CAM (S2C) framework refines Class Activation Maps (CAMs) using prototype-based contrastive learning and CAM-based prompting, improving class-specific segmentation masks~\cite{kweon2024sam}. Additionally, SAM's current robustness across segmentation tasks diminishes when applied to non-RGB data such as depth or event-based data, highlighting the need for specialized adaptations~\cite{yao2024sam}.

\subsection{Parameter-Efficient Fine-Tuning with LoRA and MoE}
Fine-tuning large pre-trained models like SAM for specific tasks often incurs high computational costs. Parameter-efficient fine-tuning (PEFT) techniques such as soft prompts, adapters, and LoRA provide efficient alternatives~\cite{han2024parameter}. LoRA introduces low-rank matrices into pre-trained models, allowing efficient adaptation by fine-tuning a minimal number of additional parameters while keeping the majority of the model weights frozen~\cite{hu2021lora}. Extensions like DyLoRA~\cite{valipour2022dylora} and SoRA~\cite{ding2023sparse} dynamically adjust the rank during training, improving adaptability across diverse tasks.

LoRA's modularity allows integration with MoE architectures, which dynamically activate specific LoRA modules based on task requirements. Routing mechanisms such as static top-k selection~\cite{wu2023mole,li2024mixlora} or dynamic thresholding~\cite{liu2024adamole,gao2024xlora} enable efficient selection of LoRA modules. Structural integrations like LoRAMoE~\cite{li2024loramoe}, which incorporates LoRA modules into feed-forward layers, and MoELoRA~\cite{pang2024moelora}, which integrates LoRA modules into both self-attention and feed-forward layers, further enhance flexibility. MixLoRA~\cite{li2024mixlora} combines LoRA modules in self-attention layers and merges them with shared feed-forward layers to optimize computational efficiency and representation learning.

Although SAM demonstrates strong generalization capabilities, it faces limitations in adapting to semantic segmentation tasks involving non-RGB modalities. Our framework represents the first attempt to adapt SAM for multi-modal semantic segmentation by leveraging an MLE tailored to specific modalities, including depth, LiDAR, and event-camera data. We propose a novel routing strategy within the MoE framework to ensure adequate cross-modal consistency, addressing the challenges inherent in multi-modal integration.
\begin{figure*}[htp]
  \begin{center}
     \includegraphics[width=\linewidth]{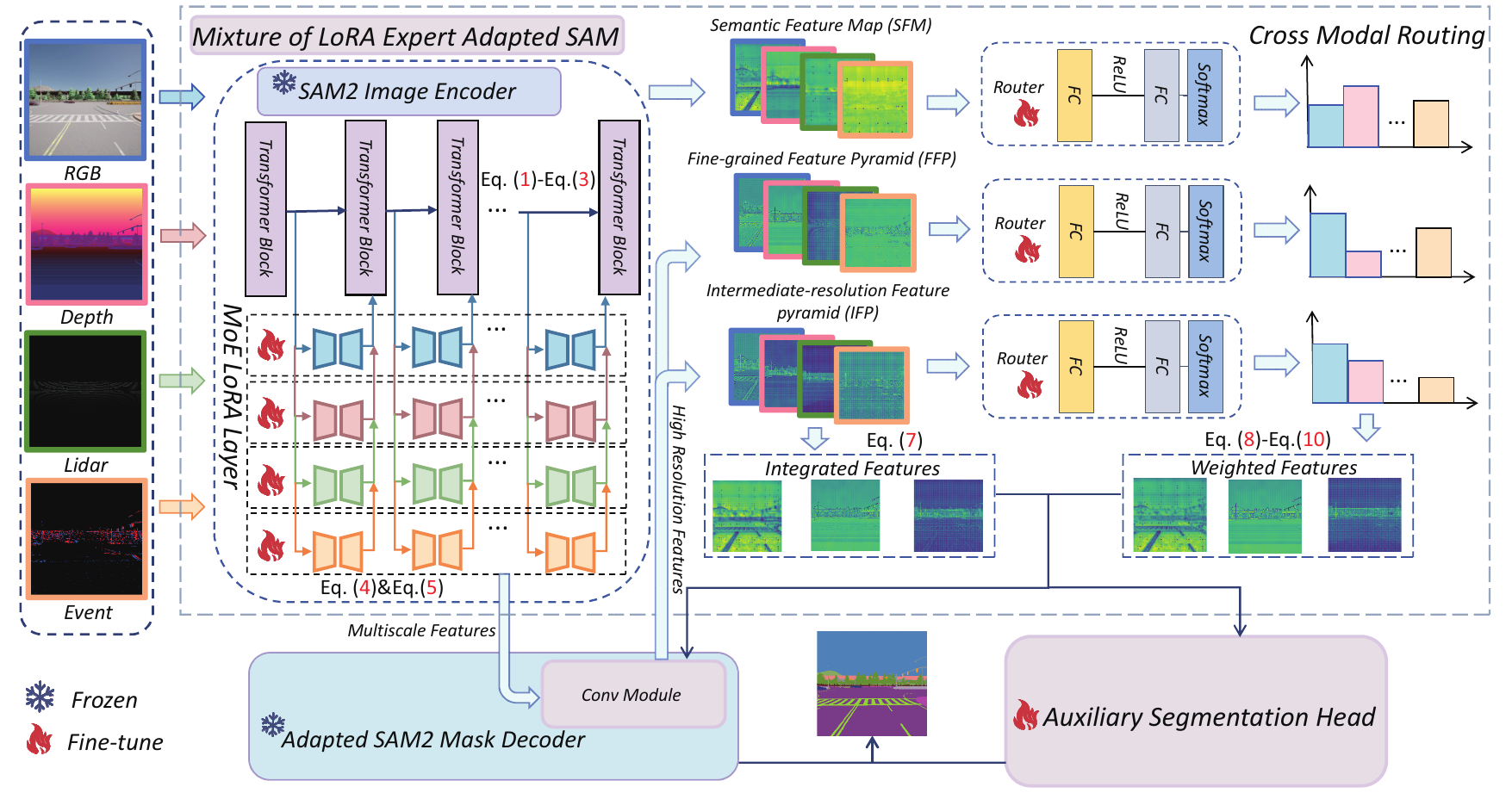}
     \caption{Illustration of the proposed \textbf{MLE-SAM} framework for multi-modal semantic segmentation. The architecture combines multi-scale features from a frozen image encoder fine-tuned with LoRA layers. Semantic feature maps and feature pyramids across modalities are averaged and refined via a top-k mechanism. Fused features are processed with a dual-pathway strategy.}\label{fig:framework}
  \end{center}
\end{figure*}

\section{Methodology}
\subsection{Preliminary}
\noindent \textbf{Segment Anything Model.} The SAM2 architecture is a transformer-based framework~\cite{li2024transformer} developed for instance segmentation, integrating three key components: a hierarchical backbone, a Feature Pyramid Network (FPN)-based neck, and a mask decoder. The hierarchical backbone adopts the Hiera architecture~\cite{ryali2023hiera} as a multi-scale feature extractor, embedding input images into high-dimensional feature spaces via a patch embedding mechanism. This backbone processes features hierarchically, doubling their dimensionality and reducing spatial resolution at each stage. These transformations leverage a combination of window-based multi-head self-attention and pooling operations, enabling the model to capture spatial and semantic relationships across varying scales. The FPN-based neck refines and consolidates these features by aligning feature dimensions from different stages, producing a unified multi-scale representation. Through its lateral connections and top-down pathways, the FPN merges fine-grained details from shallow layers with high-level semantic information from deeper layers. A sine-based positional encoding is incorporated to encode spatial relationships, enhancing the fused features for precise mask generation. The mask decoder employs transformer-based cross-attention with learnable mask tokens that iteratively interact with the fused features and positional encodings. These tokens are refined across multiple layers of cross-attention and feedforward operations. An upscaling module ensures that the final segmentation masks are high-quality and fine-grained. Moreover, the decoder's ability to output multiple masks allows it to disambiguate overlapping regions and effectively handle complex scenes.

\subsection{Framework Overview}

Building on the SAM2 framework, we propose a customized SAM2 architecture, namely \textbf{MLE-SAM} framework, designed explicitly for multi-modal semantic segmentation task, as illustrated in Figure~\ref{fig:framework}. This customization begins by freezing the pre-trained image encoder and fine-tuning it with LoRA layers, efficiently adapting the model to new visual modalities while preserving its intensive pre-trained knowledge. The image encoder processes input visual modalities $X$ to generate Semantic Feature Map (SFM) $Y_n^m$, which are further transformed by the mask decoder's convolutional module into two additional feature pyramids: a Fine-grained Feature Pyramid (FFP) $Y_0^m$ and an Intermediate-resolution Feature Pyramid (IFP) $Y_1^m$. These feature pyramids and the SFM enhance the model’s spatial and semantic representation capabilities.

To achieve an integrated feature representation, we propose a framework that combines the SFM, FFP, and IFP by averaging these representations across modalities to derive the integrated feature \( \overline{Y}_i \), where \( i \in \{0, 1, n\} \). To further refine this integration, a selective top-\(k\) mechanism is employed, generating weighted feature maps \( \hat{Y}_i \) that prioritize salient information for each index \( i \). These refined features, \( \overline{Y}_i \) and \( \hat{Y}_i \), are subsequently fused into a unified feature representation \( \tilde{Y}_i \), forming the input for downstream semantic segmentation.

The unified feature \( \tilde{Y}_i \) is processed using a dual-pathway mask prediction strategy to enhance segmentation accuracy. In the first pathway, the fused features are fed into the SAM2 mask decoder, which utilizes a frozen transformer block to extract mask tokens from the SFM. These tokens interact with the fine-grained and intermediate-resolution pyramids to construct a high-resolution feature representation. This representation is further refined by a hypernetwork to produce precise segmentation masks, denoted as \( \mathbf{\tilde{S}_0} \).

In the second pathway, the fused features are processed by an auxiliary segmentation head comprising three Multi-Layer Perceptrons (MLPs) and a series of upscaling layers. The outputs of this pathway are concatenated, passed through dropout layers to prevent overfitting, and fused linearly to predict an alternative set of high-resolution masks, \( \mathbf{\tilde{S}_1} \).
The final segmentation output is derived by combining the predictions from both pathways, leveraging their complementary strengths. This dual-pathway design effectively addresses the challenges posed by multi-modal data distributions and diverse feature scales, ensuring robust and accurate semantic segmentation across multiple modalities.

\subsection{Hierarchical Multi-Modal Feature Extraction with LoRA}
Give the input set for \( M \) modalities \( X = \{X^m \in \mathbb{R}^{H \times W \times C} \,|\, m \in [1, M]\} \), where \( H \), \( W \), and \( C \) represent the height, width, and number of channels of each modality, respectively. The index \( m \) denotes a specific modality, such as RGB, depth, LiDAR, or event camera. Each modality is processed independently through the hierarchical backbone network of Hiera to extract multi-scale features.

Initially, a patch embedding operation transforms each input \( X^m \) into an embedded feature map \( P(X^m) \in \mathbb{R}^{H_0 \times W_0 \times d} \) as shown in Eq.~(\ref{equ:patch}), where \( W_e \in \mathbb{R}^{C \times d} \) is a weight matrix, \( b_e \in \mathbb{R}^d \) is a bias vector, \( d \) is the dimensionality of the feature embedding, and \( H_0 = H / s_0 \), \( W_0 = W / s_0 \) denote the down-sampled height and width after applying a down-sampling factor \( s_0 \).
\begin{equation}\label{equ:patch}
  P(X^m) = X^m W_e + b_e
\end{equation}

The backbone of SAM2 progressively reduces spatial resolution while increasing feature dimensionality over \( n \) stages, producing multi-scale feature maps as defined in Eq.~(\ref{equ:msfeature}), where \( H_i = H / s_i \), \( W_i = W / s_i \), and \( s_i = 2^{i+2} \) defines the down-sampling factor at stage \( i \). The number of channels at stage \( i \) is denoted by \( C_i \).
\begin{equation}\label{equ:msfeature}
  \{X_i^m \in \mathbb{R}^{C_i \times H_i \times W_i} \,|\, i \in [0, n], \, m \in [1, M]\}
\end{equation}

Each stage employs window-based multi-head self-attention to extract features, as shown in Eq.~(\ref{equ:qkv}), where \( Q \), \( K \), and \( V \) are the query, key, and value matrices, \( d_k \) is the dimensionality of the key matrix, and \(\text{softmax}\) applies along the last dimension.
\begin{equation}\label{equ:qkv}
    \text{Attention}(Q, K, V) = \text{softmax}\left(\frac{Q K^\top}{\sqrt{d_k}}\right) V
\end{equation}

To enhance efficiency and modality-specific adaptation, we introduce a LoRA layer to update the query and value projections, as shown in Eq.~(\ref{equ:qv}), where \( W_a^Q, W_a^V \in \mathbb{R}^{d \times r} \) and \( W_b^Q, W_b^V \in \mathbb{R}^{r \times d} \) are low-rank matrices with \( r \ll d \) as the rank parameter. These updates yield augmented projections, as defined in Eq.~(\ref{equ:augmented_projections}). LoRA parameters are modality-specific and trained independently while freezing the backbone parameters, ensuring efficient cross-modal adaptation.
\begin{equation}\label{equ:qv}
    \Delta Q^m = W_a^Q W_b^Q, \quad \Delta V^m = W_a^V W_b^V
\end{equation}

\begin{equation}\label{equ:augmented_projections}
    Q'^m = Q^m + \Delta Q^m, \quad V'^m = V^m + \Delta V^m
\end{equation}

Hierarchical features are refined using an FPN, which integrates lateral and top-down pathways to enhance diverse multi-scale features. At each stage \( i \), the input feature map \( X_i^m \) undergoes a precise lateral convolution operation, yielding a refined modality-specific feature map \( Z_i^m \in \mathbb{R}^{d \times H_i \times W_i} \). This operation reduces the channel dimensionality to \( d \) while preserving the essential spatial dimensions \( H_i \) and \( W_i \), ensuring robust consistency in spatial resolution and compatibility for subsequent fusion operations within the FPN.

Let \( \mathcal{L}\) denote the set of layers where top-down fusion is applied. For each layer \( i \in \mathcal{L} \), top-down fusion combines feature representations from deeper layers with those at the current stage, producing the fused feature map \( Y_i^m \). This fusion process is mathematically defined in Eq.~(\ref{equ:fusedfm}).
\begin{equation}\label{equ:fusedfm}
  Y_i^m = 
\begin{cases} 
\frac{Z_i^m + \text{Upsample}(Y_{i+1}^m)}{2}, & i \in \mathcal{L} \\
Z_i^m, & i \notin \mathcal{L}.
\end{cases}
\end{equation}
Here, \( Y_i^m \in \mathbb{R}^{d \times H_i \times W_i} \) represents the fused feature map at stage \( i \), integrating modality-specific features \( Z_i^m \) with the upsampled features from the subsequent layer \( Y_{i+1}^m \). The \(\text{Upsample}\) operation adjusts the spatial resolution of \( Y_{i+1}^m \) to match that of \( Z_i^m \), ensuring accurate integration. The hierarchical refinement that underlies the multi-scale feature representation of the FPN is central to this fusion process.

\subsection{Dynamic Multi-Modal Feature Fusion with MoE and Routing Mechanisms}
The FPN is employed to generate three distinct feature maps for each modality, designed to capture semantic and spatial information at multiple different resolutions: the \textit{SFM} (\( Y_n^m \in \mathbb{R}^{d \times H_n \times W_n} \)), the \textit{FFP} (\( Y_0^m \in \mathbb{R}^{d \times H_0 \times W_0} \)), and the \textit{IFP} (\( Y_1^m \in \mathbb{R}^{d \times H_1 \times W_1} \)). To improve the overall representational capacity of the finer-resolution feature maps (\( Y_0^m \) and \( Y_1^m \)), 1x1 convolutional layers are applied to reduce their channel dimensions while preserving spatial resolution. Following these operations, the dimensions are transformed such that \( Y_0^m \in \mathbb{R}^{d / 8 \times H_0 \times W_0} \) and \( Y_1^m \in \mathbb{R}^{d / 4 \times H_1 \times W_1} \), ensuring a compact and efficient representation suitable for subsequent fusion and effective analysis.

To aggregate features across modalities, the integrated feature map \( \overline{Y}_i \) for \( i \in \{0, 1, n\} \) is computed by averaging the features across all modalities, as shown in Eq.~(\ref{equ:overliney}).
\begin{equation}\label{equ:overliney}
  \overline{Y}_i = \frac{1}{M} \sum_{m=1}^M Y_i^m, \quad i \in \{0, 1, n\}
\end{equation}
where \( Y_i^m \) denotes the feature map for modality \( m \) at pyramid level \( i \). This operation ensures uniform aggregation, capturing a holistic representation of multi-modal features. However, the equal-weight assumption in \( \overline{Y}_i \) may be suboptimal when certain modalities are more informative than others. To address this limitation, a MoE mechanism is introduced to assign dynamic weights to features based on their relevance, enabling the model to prioritize significant features while attenuating irrelevant information.

For the cross-modal routing procedure, spatially averaged embeddings \( \mathbf{f}_i^m \) are computed for each modality and feature level as compact representations of spatial information. These embeddings, defined in Eq.~(\ref{equ:fm}), are derived by averaging spatial features over height \( H_i \) and width \( W_i \). Here \( Y_i^m(h, w) \) represents the feature map of modality \( m \) at spatial location \( (h, w) \) for level \( i \).

\begin{equation}\label{equ:fm}
  \mathbf{f}_i^m = \frac{1}{H_i \cdot W_i} \sum_{h=1}^{H_i} \sum_{w=1}^{W_i} Y_i^m(h, w), \quad i \in \{0, 1, n\}
\end{equation}

Routing weights \( \mathbf{w}_i^m \), which quantify the importance of each modality for feature integration, are calculated using a linear transformation followed by an activation function \( \sigma \), as described in Eq.~(\ref{equ:wnm}), where \( \mathbf{W}_i \in \mathbb{R}^{D \times d} \) is the weight matrix, \( \mathbf{b}_i \in \mathbb{R}^D \) is the bias term, and \( \sigma \) represents a softmax function to ensure proper normalization of the routing weights.

\begin{equation}\label{equ:wnm}
    \mathbf{w}_i^m = \sigma \left( \mathbf{W}_i \cdot \mathbf{f}_i^m + \mathbf{b}_i \right), \quad i \in \{0, 1, n\}
\end{equation}

The routing mechanism dynamically selects features from the most relevant modalities based on their routing weights. For each feature level \( i \), the top-\( k \) modalities with the highest routing weights \( \mathbf{w}_i^m \) are identified. This ensures that only the most significant modalities contribute to the final feature representation. The fused feature map \( \hat{Y}_i \) is then computed as Eq.~(\ref{equ:yi}), where \( \text{Top-k} \) selects the weights corresponding to the top-\( k \) modalities, \( \odot \) denotes element-wise multiplication, and \( Y_i^m \) represents the feature map of modality \( m \) at level \( i \). 

\begin{equation}\label{equ:yi}
  \hat{Y}_i = \sum_{m=1}^M \text{Top-k}\left( \mathbf{w}_i^1, \dots, \mathbf{w}_i^M \right) \odot Y_i^m, \quad i \in \{0, 1, n\}
\end{equation}

This fusion strategy enables the model to effectively adjust the contribution of each modality, integrating both global information and modality-specific nuances into a cohesive feature representation. By prioritizing the most relevant modalities for each feature level, the approach enhances the model's capacity to handle multi-modal data and capture complementary information across modalities.

By combining \( \overline{Y} \) and \( \hat{Y} \) to the unified feature map \( \tilde{Y} \), the proposed framework effectively balances uniform aggregation for comprehensive feature representation and dynamic weighting for selective feature refinement, resulting in a robust multi-modal fusion strategy.

\subsection{Adapted Mask Decoder with Auxiliary Segmentation Head}
Next, we employ a dual-pathway mask prediction strategy on the unified feature map \( \tilde{Y} \) to generate high-resolution segmentation masks.

\begin{figure}[t!]
  \begin{center}
     \includegraphics[width=\linewidth]{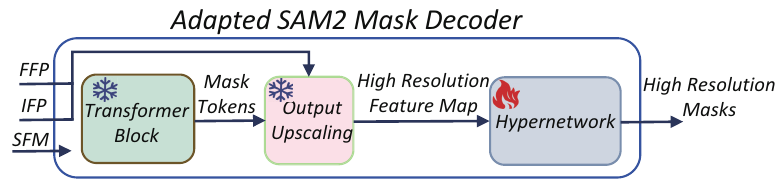}
     \caption{Hierarchical Refinement Pathway for High-Resolution Embedding}\label{fig:f1}
  \end{center}
\end{figure}

In the first pathway shown in Figure~\ref{fig:f1}, we extend SAM2's mask decoder to produce high-resolution multimasks. This involves generating high-resolution segmentation logits, denoted as \( \mathbf{\tilde{S}_0} \in \mathbb{R}^{\mathcal{C} \times H_0 \times W_0} \), through a structured multi-scale fusion process. Here, \( \mathcal{C} \) represents the number of segmentation categories. The backbone features \( \tilde{\mathbf{Y}}_n \in \mathbb{R}^{d \times H_n \times W_n} \), which encapsulate global semantic context, are processed via a transformer-based decoder \( f_{\text{dec}} \), producing low-resolution logits. These logits are iteratively refined by incorporating spatially detailed features from intermediate-resolution feature maps \( \tilde{\mathbf{Y}}_1 \in \mathbb{R}^{d / 4 \times H_1 \times W_1} \) and fine-grained feature maps \( \tilde{\mathbf{Y}}_0 \in \mathbb{R}^{d / 8 \times H_0 \times W_0} \). This hierarchical refinement process is mathematically described as Eq~(\ref{eq:combined_high_res_refined}), where \( f_{\text{dec}} \) denotes the transformer-based decoding operation applied to \( \tilde{\mathbf{Y}}_n \), \(\text{Upsample}\) performs bilinear upsampling to match spatial resolutions, and \(\text{Conv}\) is a \(1 \times 1\) convolution for channel alignment.

\begin{equation}\label{eq:combined_high_res_refined}
  \begin{aligned}
      \mathbf{S}_\text{low} &= f_{\text{dec}}(\tilde{\mathbf{Y}}_n) \\
      \mathbf{S}_\text{inter} &= \text{Upsample}(\mathbf{S}_\text{low}) + \text{Conv}(\tilde{\mathbf{Y}}_1) \\
      \mathbf{\tilde{S}_0} &= \text{Upsample}(\mathbf{S}_\text{inter}) + \text{Conv}(\tilde{\mathbf{Y}}_0)
  \end{aligned}
\end{equation}

\begin{figure}[t!]
  \begin{center}
     \includegraphics[width=\linewidth]{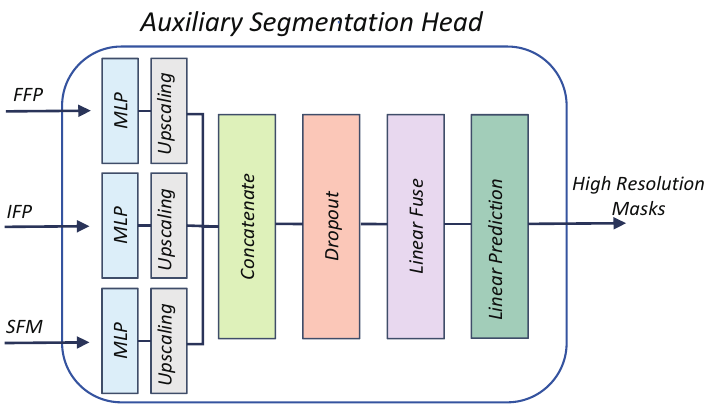}
     \caption{Multi-Scale Feature Fusion Pathway for High-Resolution Embedding}\label{fig:f2}
  \end{center}
\end{figure}

As shown in Figure~\ref{fig:f2}, the second pathway utilizes a feature fusion mechanism to integrate multi-scale features into a unified high-resolution embedding. Specifically, backbone features \( \tilde{\mathbf{Y}}_n \in \mathbb{R}^{d \times H_n \times W_n} \), \( \tilde{\mathbf{Y}}_1 \in \mathbb{R}^{d / 4 \times H_1 \times W_1} \), and \( \tilde{\mathbf{Y}}_0 \in \mathbb{R}^{d / 8 \times H_0 \times W_0} \) are first transformed via MLPs and upsampled to a common target resolution \( H_t \times W_t \) using bilinear interpolation. This results in upsampled feature maps \( \mathbf{Y}_n^\text{up} \in \mathbb{R}^{d / 8 \times H_t \times W_t} \), \( \mathbf{Y}_1^\text{up} \in \mathbb{R}^{d / 8 \times H_t \times W_t} \), and \( \mathbf{Y}_0^\text{up} \in \mathbb{R}^{d / 8 \times H_t \times W_t} \), respectively. These upsampled features are then concatenated along the channel dimension and passed through a linear fusion layer \( f_{\text{fuse}} \), followed by a prediction layer \( f_{\text{pred}} \), to produce the high-resolution segmentation logits \( \mathbf{\tilde{S}_1} \) as described in Eq.~(\ref{eq:final_high_res_fusion}). \( f_{\text{fuse}} \) effectively integrates features from multiple scales, while \( f_{\text{pred}} \) generates the segmentation logits. This dual-pathway approach captures both global and local contextual information, thereby enhancing segmentation accuracy and robustness.

\begin{equation}\label{eq:final_high_res_fusion}
  \mathbf{\tilde{S}_1} = f_{\text{pred}} \left( f_{\text{fuse}} \left( \text{Concat} \left( 
      \mathbf{Y}_n^\text{up}, \mathbf{Y}_1^\text{up}, \mathbf{Y}_0^\text{up} \right) \right) \right)
\end{equation}

The training process minimizes a loss function that integrates the Online Hard Example Mining Cross-Entropy (OhemCrossEntropy) loss~\cite{shrivastava2016training}, which focuses on hard-to-predict pixels to improve model robustness and efficiency. The ground truth segmentation labels \( \mathbf{L} \in \mathbb{R}^{H_t \times W_t} \) are defined such that \(\mathbf{L}(i,j) \in \{0, 1, \dots, \mathcal{C}-1, 255\}\), where 255 indicates the ignore label. The OhemCrossEntropy loss for a single prediction map \( \mathbf{\tilde{S}} \) is given by Eq.~(\ref{equ:ohem}).
\begin{equation}\label{equ:ohem}
  \mathcal{L}_{\text{Ohem}}(\mathbf{\tilde{S}}, \mathbf{L}) = \frac{1}{n_{\text{min}}} \sum_{i \in \mathcal{H}} \mathcal{L}_{\text{CE}}(\mathbf{\tilde{S}}(i), \mathbf{L}(i))
\end{equation}

where \(\mathcal{L}_{\text{CE}}\) is the pixel-wise cross-entropy loss, and \(\mathcal{H}\) represents the set of hardest pixels, selected based on prediction difficulty. The normalization factor \(n_{\text{min}} = \max(|\mathcal{H}|, n_{\text{threshold}})\) ensures that a sufficient number of complex examples are included, where \(n_{\text{threshold}} = n_{\text{total}} / 16\), and \(n_{\text{total}}\) is the total number of valid pixels in the image.

The overall loss function incorporates the OhemCrossEntropy loss applied to both \( \mathbf{\tilde{S}_0} \) and \( \mathbf{\tilde{S}_1} \), as defined in Eq.~(\ref{equ:loss}).

\begin{equation}\label{equ:loss}
  \mathcal{L} = w_0 \cdot \mathcal{L}_{\text{Ohem}}(\mathbf{\tilde{S}_0}, \mathbf{L}) + w_1 \cdot \mathcal{L}_{\text{Ohem}}(\mathbf{\tilde{S}_1}, \mathbf{L})
\end{equation}

where \( w_0, w_1 \in \mathbb{R}^+ \) are scalar weights that control the relative importance of each loss term.

\section{Experiments}
\subsection{Experimental Setup}

\noindent \textbf{Datasets.} 
To comprehensively evaluate the performance of the proposed MLE-SAM model in multi-modal semantic segmentation, three distinct datasets were selected, each targeting specific challenges in autonomous driving and material segmentation tasks. These datasets provide complementary benchmarks to address real-world complexities such as adverse weather conditions, sensor failures, and multi-modal fusion in diverse scenarios.

\noindent \textbf{The DELIVER dataset}~\cite{zhang2023delivering} is a large-scale multi-modal benchmark designed explicitly for semantic segmentation in autonomous driving scenarios. Developed using the CARLA simulator, it incorporates data from four modalities:RGB (R), Depth (D), LiDAR (L) and Event (E), enabling advanced multi-modal fusion research. The dataset consists of 7,885 front-view images, each with a resolution of 1,042 by 1,042 pixels, partitioned into 3,983 images for training, 2,005 for validation, and 1,897 for testing. Semantic segmentation is supported across 25 distinct classes, with each data sample providing six panoramic views covering a field of view of \(91^\circ \times 91^\circ\). To emulate real-world challenges, DELIVER introduces four adverse weather conditions and five sensor failure cases, including motion blur, overexposure, and LiDAR jitter. 

\noindent \textbf{The MUSES dataset}~\cite{brodermann2024muses} is a multi-modal benchmark tailored for dense semantic perception in autonomous driving under challenging environmental conditions like rain, snow, fog, and nighttime. It provides 2,500 samples with high-quality 2D panoptic annotations spanning 19 semantic classes. The dataset is divided into 1,500 training samples, 250 validation samples, and 750 test samples, each captured at a resolution of 1,920 by 1,080 pixels. MUSES integrates synchronized data from three modalities: a frame camera (F), an event camera (E), and a LiDAR (L), offering diverse inputs for tasks including semantic segmentation, panoptic segmentation, and uncertainty-aware panoptic segmentation.

\noindent \textbf{The MCubeS dataset}~\cite{liang2022multimodal} is a multi-modal benchmark designed for material semantic segmentation, focusing on dense per-pixel recognition of material categories in challenging outdoor scenes. It includes 500 annotated image sets capturing 42 scenes with four distinct imaging modalities: RGB, near-infrared (NIR), and polarization represented by the Angle of Linear Polarization (AoLP) and the Degree of Linear Polarization (DoLP). The dataset is divided into 302 images for training, 96 for validation, and 102 for testing, with each image at a resolution of high-quality 1920 by 1080 pixels. It annotates 20 material classes, including asphalt, concrete, metal, fabric, water, and grass types.

\noindent \textbf{Multi-modal Segmentation Evaluation.}
We evaluated the proposed MLE-SAM method for multi-modal semantic segmentation against three state-of-the-art approaches, namely CMNeXt~\cite{zhang2023delivering}, CWSAM~\cite{pu2024classwise}, and SAM-LoRA, across three benchmark datasets. For fairness comparison, the backbone architectures were standardized as follows: MiT-B0 was employed for CMNeXt, ViT-B served as the backbone for both CWSAM and SAM-LoRA, while MLE-SAM utilized Hiera-B+ as its backbone. Detailed implementation details is provided in Appendix~\ref{appendix:a}. The evaluation included various combinations of input modalities to assess each method's ability to integrate and utilize multi-modal information. Additionally, quantitative analysis was conducted on the DELIVER dataset, comparing trainable parameters and performance under challenging environmental conditions such as cloudy, foggy, motion blur, overexposure, underexposure, LiDAR jitter, and event low resolution. This systematic assessment provides a comprehensive understanding of each method's robustness and efficiency across diverse scenarios.
\begin{table}[t!] 
  \centering
  \caption{Experimental comparison on DELIVER across various modality combinations.}\label{tab:deliver}
  \begin{tabular}{ccccc}
  \midrule
  \textbf{Method} & \textbf{Modal} & \textbf{Backbone} & \textbf{mIoU} & $\boldsymbol{\triangle\uparrow}$      \\ \midrule
  CMNeXt~\cite{zhang2023delivering}          & RGB            & MiT-B0            & 51.29         & -     \\
  CWSAM~\cite{pu2024classwise}           & RGB            & ViT-B             & 51.59         & 0.30   \\
  SAM-LoRA        & RGB            & ViT-B             & \underline{51.84}         & 0.55  \\
\rowcolor{gray!10}  MLE-SAM        & RGB            & Hiera-B+   & \textbf{55.23}         & 3.94  \\ \midrule
  CMNeXt~\cite{zhang2023delivering}          & RGB-Depth      & MiT-B0            & 59.61         & -     \\
  CWSAM~\cite{pu2024classwise}           & RGB-Depth      & ViT-B             & 58.64         & -0.97 \\
  SAM-LoRA        & RGB-Depth      & ViT-B             & \underline{60.25}         & 0.64  \\
\rowcolor{gray!10}  MLE-SAM        & RGB-Depth      & Hiera-B+   & \textbf{63.57}         & 3.96  \\ \midrule
  CMNeXt~\cite{zhang2023delivering}          & RGB-D-Event    & MiT-B0            & 59.84         & -     \\
  CWSAM~\cite{pu2024classwise}           & RGB-D-Event    & ViT-B             & 56.22         & -3.62 \\
  SAM-LoRA        & RGB-D-Event    & ViT-B             & \underline{60.08}         & 0.24  \\
\rowcolor{gray!10} MLE-SAM        & RGB-D-Event    & Hiera-B+   & \textbf{62.69}         & 2.85  \\ \midrule
  CMNeXt~\cite{zhang2023delivering}          & RGB-D-E-LiDAR  & MiT-B0            & 59.18         & -     \\
  CWSAM~\cite{pu2024classwise}           & RGB-D-E-LiDAR  & ViT-B             & 55.43         & -3.75 \\
  SAM-LoRA        & RGB-D-E-LiDAR  & ViT-B             & \underline{59.54}         & 0.36  \\
 \rowcolor{gray!10} MLE-SAM        & RGB-D-E-LiDAR  & Hiera-B+   & \textbf{64.08}         & 4.90   \\ \midrule
  \end{tabular}
  \end{table}
  
\noindent \textbf{Segmentation Evaluation with Missing Modalities and Noise.}
We then evaluated the robustness of semantic segmentation models trained with all available modalities but tested under various combinations of individual and partial modalities on DELIVER and MUSES datasets. The robustness of MLE-SAM is analyzed under Gaussian and Random noise applied to different modalities, with mean Intersection over Union (mIoU) as the primary evaluation metric.
We implemented a noise augmentation module to simulate adverse conditions for injecting Gaussian or random noise into specified modalities. Gaussian noise was generated using a standard normal distribution scaled by 50.0, while random noise was uniformly sampled within the range [-100, 100]. The noise was directly added to the image data of the targeted modality, followed by clipping pixel values to the range [0, 255] to ensure validity and prevent overflow or underflow in pixel intensities.

\subsection{Multi-modal Segmentation Comparison}

The performance comparison in Table~\ref{tab:deliver} demonstrates the efficacy of the proposed MLE-SAM model, a SAM-based approach, in semantic segmentation tasks on the DELIVER dataset. Across all tested modality combinations, MLE-SAM consistently achieves the highest mIoU scores, significantly surpassing the performance of competing methods. For the single-modality RGB configuration, MLE-SAM achieves an mIoU of 55.23\%, outperforming CMNeXt and SAM-LoRA by margins of 3.94\% and 3.39\%, respectively. When utilizing RGB and Depth modalities, the mIoU increases to 63.57\%, a gain of 3.96\% over CMNeXt and 3.32\% over SAM-LoRA. Incorporating Event data alongside RGB and Depth yields an mIoU of 62.69\%, with improvements of 2.85\% and 2.61\% over CMNeXt and SAM-LoRA, respectively. The addition of all four modalities results in the best performance for MLE-SAM, achieving an mIoU of 64.08\%, exceeding SAM-LoRA by 4.54\% and CMNeXt by 4.90\%. These results highlight the ability of MLE-SAM to effectively integrate multi-modal information, with performance gains becoming more pronounced as additional modalities are incorporated. Notably, the inclusion of all modalities leads to an mIoU improvement of 8.85\% over the RGB-only configuration, underscoring the significant advantage of multi-modal fusion in semantic segmentation.

The results in Table~\ref{tab:muses} further validate the superiority of MLE-SAM on the MUSES dataset. The model consistently achieves the highest mIoU scores across all modality combinations, significantly outperforming other methods. For single-modality Frame-camera inputs, MLE-SAM attains an mIoU of 73.95\%, surpassing CMNeXt by 30.58\% and SAM-LoRA by 8.04\%. With the Frame-camera and Event modality combination, the mIoU improves to 74.73\%, exceeding CMNeXt and SAM-LoRA by 31.34\% and 6.77\%, respectively. Adding LiDAR to Frame-camera further enhances the mIoU to 75.42\%, representing a 28.39\% improvement over CMNeXt and a 5.08\% improvement over SAM-LoRA. The integration of Frame-camera, Event, and LiDAR modalities achieves an mIoU of 74.8\%, maintaining MLE-SAM’s superior performance with gains of 28.14\% and 4.72\% over CMNeXt and SAM-LoRA, respectively. These findings highlight the robust capacity of MLE-SAM to leverage real-world multi-modal data effectively, enabling significant segmentation performance enhancements.

The results on both datasets reveal important insights into the relationship between dataset characteristics and model performance. While MLE-SAM demonstrates strong segmentation capabilities on both datasets, its higher performance on MUSES can be attributed to the alignment between the SAM pretraining corpus and the real-world nature of MUSES. As SAM-based models are pre-trained on diverse real-world images, they are inherently better suited to datasets like MUSES, which capture complex, realistic environmental conditions. Conversely, the simulated nature of the DELIVER dataset limits the full exploitation of SAM’s pre-trained knowledge.
\begin{table}[t!] 
  \centering
  \caption{Experimental results on the MUSES.}\label{tab:muses}
  \begin{tabular}{ccccc}
  \midrule
  \textbf{Method} & \textbf{Modal} & \textbf{Backbone} & \textbf{mIoU} &  $\boldsymbol{\triangle\uparrow}$     \\ \midrule
  CMNeXt~\cite{zhang2023delivering}          & Frame            & MiT-B0            & 43.37          & -     \\
  CWSAM~\cite{pu2024classwise}           & Frame            & ViT-B             & 55.41          & 12.04 \\
  SAM-LoRA        & Frame            & ViT-B             & \underline{65.91}          & 22.54 \\
\rowcolor{gray!10} MLE-SAM        & Frame            & Hiera-B+   & \textbf{73.95}          & 30.58 \\ \midrule
  CMNeXt~\cite{zhang2023delivering}          & Frame-Event      & MiT-B0            & 43.39          & -     \\
  CWSAM~\cite{pu2024classwise}           & Frame-Event      & ViT-B             & 41.77          & -1.62 \\
  SAM-LoRA        & Frame-Event      & ViT-B             & \underline{67.96}          & 24.57 \\
\rowcolor{gray!10}  MLE-SAM        & Frame-Event      & Hiera-B+   & \textbf{74.73}          & 31.34 \\ \midrule
  CMNeXt~\cite{zhang2023delivering}          & Frame-LiDAR      & MiT-B0            & 47.03          & -     \\
  CWSAM~\cite{pu2024classwise}           & Frame-LiDAR      & ViT-B             & 40.69          & -6.34 \\
  SAM-LoRA        & Frame-LiDAR      & ViT-B             & \underline{70.34}          & 23.31 \\
\rowcolor{gray!10}  MLE-SAM        & Frame-LiDAR      & Hiera-B+   & \textbf{75.42}          & 28.39 \\ \midrule
  CMNeXt~\cite{zhang2023delivering}          & Frame-E-LiDAR    & MiT-B0            & 46.66          & -     \\
  CWSAM~\cite{pu2024classwise}           & Frame-E-LiDAR    & ViT-B             & 49.98          & 3.32  \\
  SAM-LoRA        & Frame-E-LiDAR    & ViT-B             & \underline{70.08}          & 23.42 \\
\rowcolor{gray!10}  MLE-SAM        & Frame-E-LiDAR    & Hiera-B+   & \textbf{74.8}           & 28.14 \\ \midrule
  \end{tabular}
  \end{table}

  Table~\ref{tab:mcubes} showcases MLE-SAM's performance on the MCubeS dataset, further affirming its capability for multi-modal semantic segmentation. With the RGB-AOLP modality combination, MLE-SAM achieves an mIoU of 50.61\%, outperforming SAM-LoRA by 1.87\%, CWSAM by 0.83\%, and CMNeXt by a significant 13.40\%. The inclusion of DoLP alongside RGB and AOLP raises the mIoU to 50.89\%, surpassing SAM-LoRA by 1.54\%, CWSAM by 2.62\%, and CMNeXt by 12.17\%. Adding NIR to the RGB-AOLP-DoLP configuration achieves the highest mIoU of 51.02\%, with respective improvements of 1.56\% over SAM-LoRA, 0.43\% over CWSAM, and a remarkable 14.86\% over CMNeXt. These results underscore MLE-SAM’s proficiency in integrating multi-modal information for dense per-pixel material segmentation, particularly in challenging outdoor scenes.

 In summary, the experimental results across the DELIVER, MUSES, and MCubeS datasets consistently demonstrate the superior performance of MLE-SAM in leveraging multi-modal data for semantic segmentation. The model achieves substantial gains over state-of-the-art competitors by utilizing complementary information from multiple modalities. Moreover, the observed performance trends highlight the importance of dataset characteristics, with real-world datasets providing more opportunities for SAM-based models to exploit their pretraining strengths fully. The consistent improvements across diverse configurations underscore MLE-SAM’s robustness and scalability, establishing it as a robust framework for advancing multi-modal segmentation tasks.
\begin{table}[t!] 
  \centering
  \caption{Experimental results on MCubeS.}\label{tab:mcubes}
  \begin{tabular}{ccccc}
  \midrule
  \textbf{Method} & \textbf{Modal} & \textbf{Backbone} & \textbf{mIoU} & $\boldsymbol{\triangle\uparrow}$  \\ \midrule
  CMNeXt~\cite{zhang2023delivering}          & RGB-AOLP       & MiT-B0            &  37.21             & - \\
  CWSAM~\cite{pu2024classwise}           & RGB-AOLP       & ViT-B             & \underline{49.78}         & 12.57 \\
  SAM-LoRA        & RGB-AOLP       & ViT-B             & 48.74         & 11.53 \\
\rowcolor{gray!10}  MLE-SAM        & RGB-AOLP       & Hiera-B+   & \textbf{50.61}         & 13.40 \\ \midrule
  CMNeXt~\cite{zhang2023delivering}          & RGB-A-DOLP     & MiT-B0            &  38.72             &  -\\
  CWSAM~\cite{pu2024classwise}           & RGB-A-DOLP     & ViT-B             & 48.27         & 9.55 \\
  SAM-LoRA        & RGB-A-DOLP     & ViT-B             & \underline{49.35}         & 10.63 \\
\rowcolor{gray!10}  MLE-SAM        & RGB-A-DOLP     & Hiera-B+   & \textbf{50.89}         & 12.17  \\ \midrule
  CMNeXt~\cite{zhang2023delivering}          & RGB-A-D-NIR    & MiT-B0            &   36.16            & - \\
  CWSAM~\cite{pu2024classwise}           & RGB-A-D-NIR    & ViT-B             & \underline{50.59}         & 14.43 \\
  SAM-LoRA        & RGB-A-D-NIR    & ViT-B             & 49.46         & 13.30 \\
\rowcolor{gray!10}  MLE-SAM        & RGB-A-D-NIR    & Hiera-B+   & \textbf{51.02}         & 14.86  \\ \midrule
  \end{tabular}
  \end{table}

\subsection{Ablation Studies and Qualitative Analysis}
The quantitative evaluation of modality combinations on the DELIVER reveals the relationship between trainable parameters and performance under various conditions. As shown in Table~\ref{tab:adverse}, under normal conditions (cloudy, foggy, and sunny), RGB-D performs best with mIoU values of 66.21\%, 63.89\%, and 65.58\%, respectively. Combining RGB and Depth enhances feature richness and robustness. Under adverse conditions (night and rainy), RGB-D-E and RGB-D-E-L outperform, with mIoU values of 60.82\% and 62.68\% for night, and 62.01\% and 62.71\% for rainy conditions. Including sparse modalities like Event and LiDAR compensates for the limitations of dense sensors in low-light and high-reflection environments by capturing high-dynamic-range data.

\begin{table*}[htp]
  \centering
  \caption{Quantitative evaluation of different modality combinations trained on DELIVER, detailing the number of trainable parameters and performance under various environmental conditions (e.g., cloudy, foggy, night, rainy, sunny, motion blur (MB), overexposure (OE), underexposure (UE), LiDAR jitter (LJ), and event low resolution (EL))}\label{tab:adverse}
  \begin{tabular}{@{}llccccccccccc@{}}
  \toprule
  \textbf{Modality} & \textbf{\#Params(M)} & \textbf{Cloudy} & \textbf{Foggy} & \textbf{Night} & \textbf{Rainy} & \textbf{Sunny} & \textbf{MB} & \textbf{OE} & \textbf{UE} & \textbf{LJ} & \textbf{EL} & \textbf{Mean} \\ \midrule
  RGB                     & 5.2                  & 58.25           & 56.07          & 47.81          & 54.67          & 58.46          & 56.95       & 49.16       & 35.65       & 54.09       & 54.69       & 55.23         \\
  Depth                   & 5.2                  & 54.25           & 54.23          & 53.31          & 51.02          & 54.17          & 52.93       & 55.17       & 53.45       & 53.95       & 50.79       & 53.72         \\
  Event                   & 5.2                  & 30.73           & 18.88          & 30.46          & 27.75          & 26.53          & 26.87       & 24.61       & 27          & 30.49       & 21.25       & 26.7          \\
  LiDAR                   & 5.2                  & 26.76           & 28.21          & 25.98          & 27             & 28.36          & 26.22       & 27.19       & 29.95       & 21.03       & 28.43       & 27.46         \\
  RGB-D                   & 10.4                 & \textbf{66.21}           & \textbf{63.89}          & \underline{62.16}          & 61.23          & \textbf{65.58}          & \textbf{63.03}       & 63.17       & \underline{57.82}       & \textbf{63.46}       & \underline{63.73}       & \underline{63.57}         \\
  RGB-D-E                 & 15.6                 & \underline{65.09}           & 61.41          & 60.82          & \underline{62.01}          & \underline{65.21}          & 62.26       & \underline{63.41}       & 56.9        & 61.32       & 62.19       & 62.69         \\
  RGB-D-E-L               & 20.79                & 64.72           & \underline{62.87}          & \textbf{62.68}          & \textbf{62.71}          & 65.4           & \underline{62.66}       & \textbf{64.28}       & \textbf{59.35}       & \underline{63.22}       & \textbf{64.15}       & \textbf{64.08}         \\ \bottomrule
  \end{tabular}
  \end{table*}
RGB-D is most effective in handling motion blur in sensor failure scenarios, achieving an mIoU of 63.03\% by leveraging complementary spatial and depth information. For more challenging conditions like overexposure, LiDAR jitter, and event low resolution, RGB-D-E-L offers the highest robustness, with mIoU values of 64.28\%, 63.22\%, and 64.15\%, respectively. This improvement comes from combining dense modalities (RGB and Depth) with sparse modalities (Event and LiDAR), where sparse data enhances performance in conditions that limit dense sensors.

From a computational perspective, trainable parameters increase from 5.2 million for single modalities like RGB or Depth to 20.79 million for the RGB-D-E-L combination. Dense sensors excel in capturing detailed information but are sensitive to noise in extreme conditions. In contrast, sparse data from Event and LiDAR improves robustness by highlighting critical features in degraded scenarios. This analysis emphasizes the importance of multi-modal fusion in enhancing robustness and adaptability, balancing dense and sparse data to ensure consistent performance across diverse environments.

  \begin{table}[t!]
    \centering
    \caption{Ablation study on DELIVER using R-D-L-E modalities, analyzing the impact of integrated features, weighted features, and an auxiliary segmentation head on the number of parameters and mIoU scores.}\label{tab:ablation}
    \begin{tabular}{ccccc}
      \midrule
      \textbf{\begin{tabular}[c]{@{}c@{}}Integrated\\ Features\end{tabular}} & \textbf{\begin{tabular}[c]{@{}c@{}}Weighted\\ Features\end{tabular}} & \textbf{\begin{tabular}[c]{@{}c@{}}Auxiliary\\ Segmentation\\ Head\end{tabular}} & \textbf{\#Params} & \textbf{mIoU} \\ \midrule
      \checkmark                                         &                                                                          &                                                                                  & 20.62             & 61.87         \\
      \checkmark                                         &                                                                          & \checkmark                                                        & 20.64             & \underline{62.03}         \\
                                                                        & \checkmark                                                &                                                                                  & 20.77             & 58.35         \\
                                                                        & \checkmark                                                & \checkmark                                                        & 20.79             & 57.99         \\
      \checkmark                                         & \checkmark                                                & \checkmark                                                        & 20.79             & \textbf{64.08}         \\ \midrule
      \end{tabular}
    \end{table}
    \begin{figure}[t!] 
      \begin{center}
         \includegraphics[width=\linewidth]{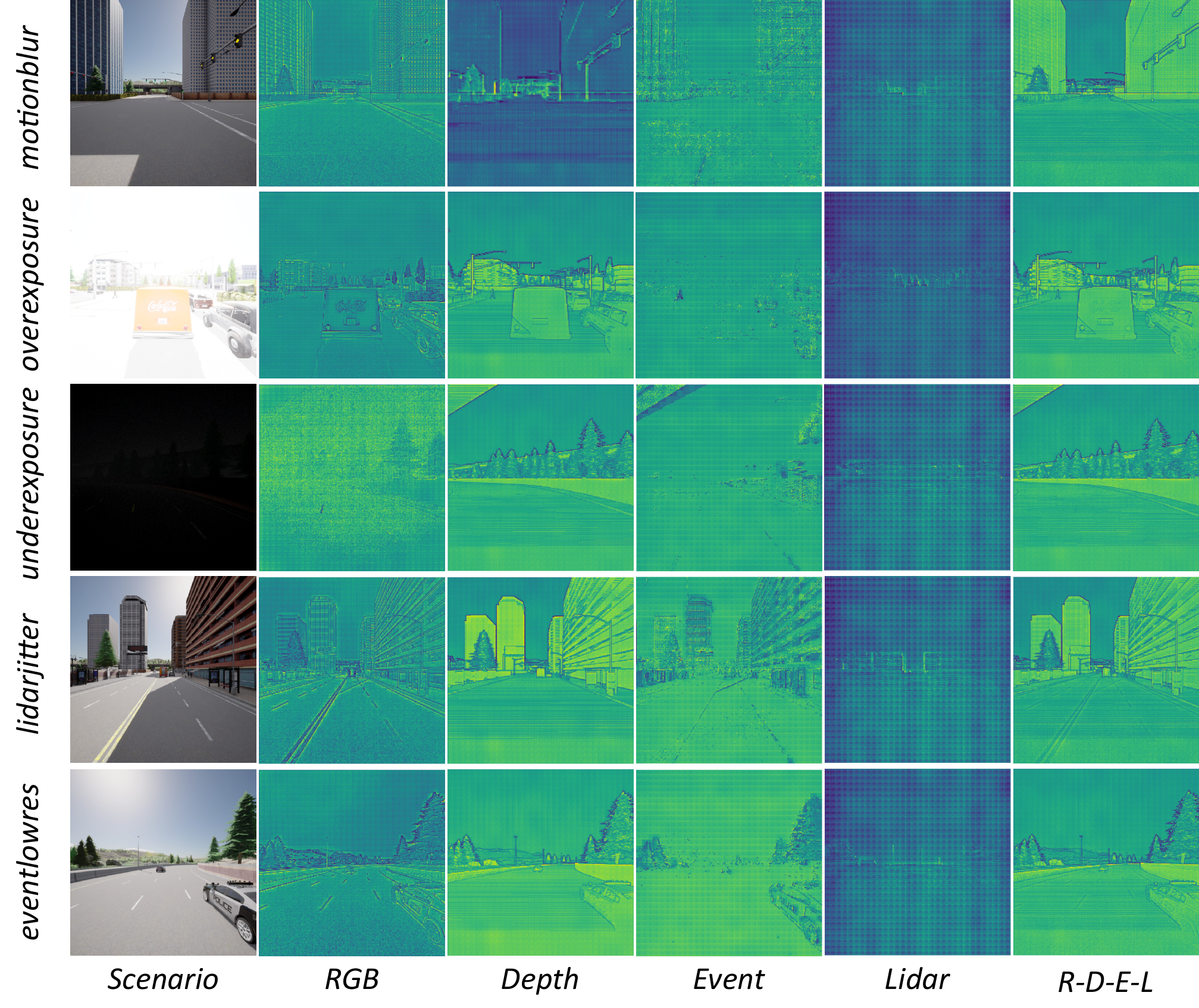}
         \caption{Visualization of extracted feature maps of DELIVER under sensor failure cases for RGB, Depth, Event, LiDAR, and R-D-E-L modalities}\label{fig:ablation1}
      \end{center}
    \end{figure}
    
Table~\ref{tab:ablation} evaluates the impact of integrated features \( \overline{Y} \), weighted features \( \hat{Y} \), and the auxiliary segmentation head on multi-modal semantic segmentation using the DELIVER with R-D-L-E modalities. The integration of \( \overline{Y} \) results in a substantial improvement in segmentation performance, achieving an mIoU of 61.87\% with 20.62 million parameters. Adding an auxiliary segmentation head with integrated features raises the mIoU to 62.03\%, with a slight parameter increase (20.64 million). In contrast, the use of weighted features \( \hat{Y} \) alone leads to inferior results, with mIoU scores of 58.35\% and 57.99\% when the auxiliary head is excluded and included, both requiring more parameters (20.77 and 20.79 million). The combination of \( \overline{Y} \) and \( \hat{Y} \), along with the auxiliary segmentation head, achieves the highest performance, with an mIoU of 64.08\% and 20.79 million parameters. These results highlight the importance of combining both feature types, as their integration enhances feature representation and segmentation accuracy.

Figure~\ref{fig:ablation1} shows the extracted feature maps under adverse sensor conditions across various modalities. The performance of each modality is affected by its intrinsic characteristics, especially in challenging environments. For example, RGB features are sensitive to lighting changes, suffering significant degradation under overexposure or underexposure. Depth and LiDAR features are vulnerable to environmental disturbances like LiDAR jitter, which introduces noise in depth estimation and spatial measurements. In contrast, combining modalities enhances robustness by leveraging complementary strengths and mitigating the limitations of individual features.

For instance, in overexposure or underexposure conditions, depth features help capture detailed object information (\textit{e.g.}, trees and cars), compensating for RGB’s underperformance. Similarly, in the presence of LiDAR jitter, combining RGB and event features improves texture representation, preserving details like building structures. These results demonstrate the effectiveness of multi-modal fusion in creating more resilient feature representations under adverse conditions.

\begin{figure*}[htp]
  \begin{center}
     \includegraphics[width=0.9\linewidth]{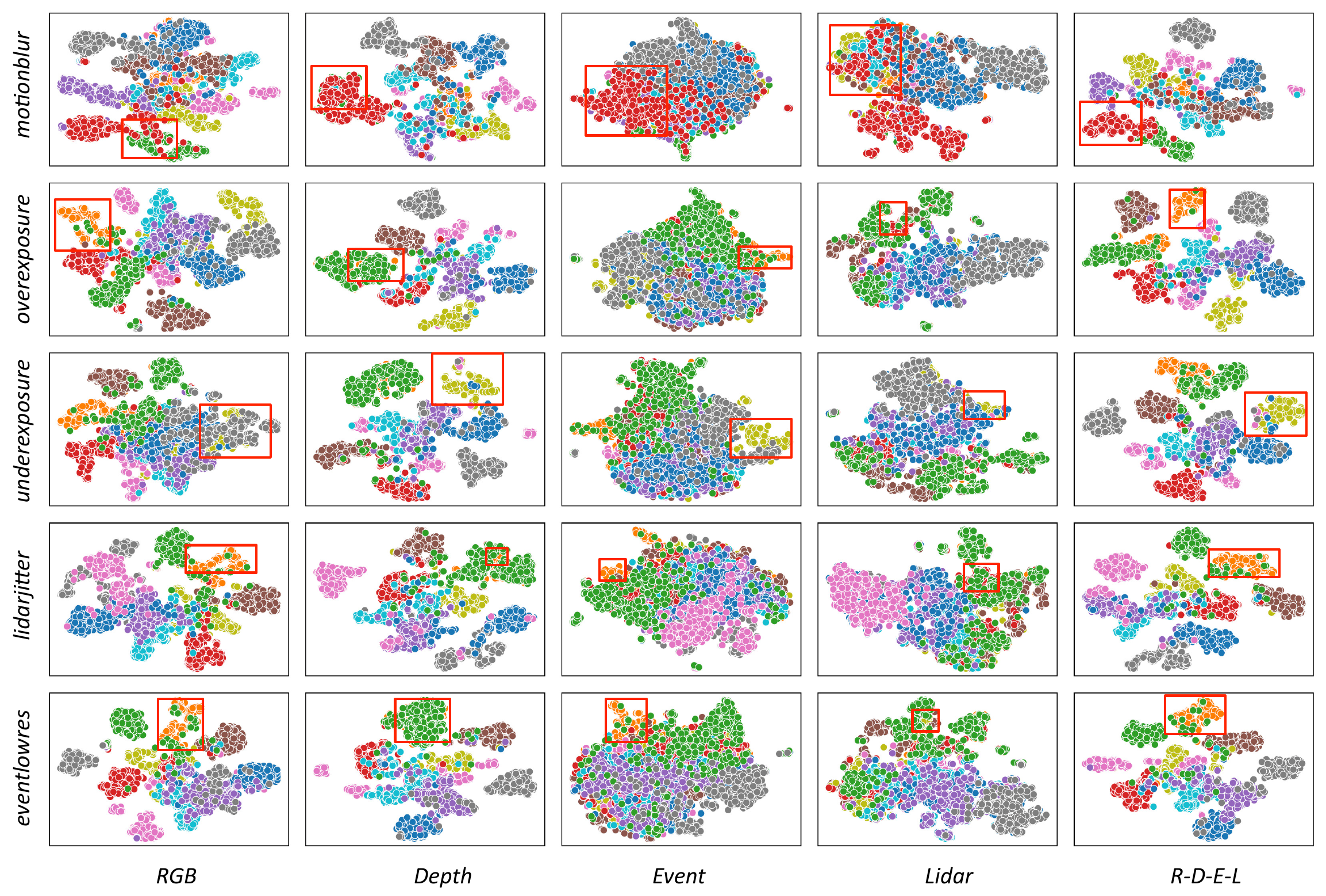}
     \caption{t-SNE visualization of pixel-level features from selected semantic classes under sensor failure scenarios in the DELIVER dataset. Each point represents a pixel, color-coded by class.}\label{fig:tsne}
  \end{center}
\end{figure*}

Figure~\ref{fig:tsne} presents the t-SNE visualizations of pixel-level features from selected semantic classes under sensor failure scenarios, highlighting substantial variations in feature separability across modalities and failure conditions. Each point in the visualization corresponds to a pixel, color-coded by its semantic class, illustrating the underlying distribution of features in the high-dimensional space. In single-modality scenarios, sensor failures result in significant class overlap, reflecting a diminished discriminative capacity of the feature representations. Conversely, multi-modal training substantially improves feature separability, demonstrating the effectiveness of multi-modal fusion in constructing robust feature representations. Notably, dense modalities, such as RGB and depth, exhibit superior class separability compared to sparse modalities like event and LiDAR, underscoring the critical role of data density in preserving semantic integrity under adverse conditions. These results emphasize the potential of multi-modal approaches to enhance semantic segmentation performance, particularly in sensor-degraded environments.

\begin{table*}[]
  \centering
  \caption{Experimental results on different modality combinations and tested under various individual and combined modality scenarios using the DELIVER dataset. The modalities include RGB (R), Depth (D), Event (E), and LiDAR (L)}\label{tab:delivermissing}
\renewcommand{\tabcolsep}{4pt}
  \begin{adjustbox}{width=\textwidth}
    \begin{tabular}{c|c|ccccccccccccccc|c|c}
      \midrule
      \multirow{2}{*}{\textbf{Method}} & \multirow{2}{*}{\textbf{Training}} & \multicolumn{15}{c|}{\textbf{DELIVER dataset}}                                                                                                                                                                                     & \multirow{2}{*}{\textbf{Mean}} & \multirow{2}{*}{$\boldsymbol{\triangle\uparrow}$}         \\ \cmidrule{3-17}
                                       &                                    & \textbf{R} & \textbf{D} & \textbf{E} & \textbf{L} & \textbf{R-D} & \textbf{R-E} & \textbf{R-L} & \textbf{D-E} & \textbf{D-L} & \textbf{E-L} & \textbf{R-D-E} & \textbf{R-D-L} & \textbf{R-E-L} & \textbf{D-E-L} & \textbf{R-D-E-L} &                                &                           \\ \midrule
      CMNeXt                           & \multirow{4}{*}{R-D-E}             & 2.69       & 0.21       & 0.78       & -          & 48.04        & 6.92         & -            & 6.92         & -            & -            & 59.84          & -              & -              & -              & -                & 17.91                          & -                         \\
      CWSAM                            &                                    & 12.3       & 35.42      & \textbf{8.16}       & -          & 27.26        & 17.44        & -            & 40.96        & -            & -            & 56.22          & -              & -              & -              & -                & 28.25                          & 9.64                      \\
      SAM-LoRA                        &                                    & \underline{18.34}      & \textbf{48.94}      & 3.36       & -          & \underline{60.08}        & \underline{18.34}        & -            & \underline{48.94}        & -            & -            & \underline{60.08}          & -              & -              & -              & -                & \underline{36.87}                          & 18.95                     \\
\rowcolor{gray!10}      MLE-SAM                             &                                    & \textbf{20.77}      & \underline{48.59}      & \underline{4.68}       & -          & \textbf{62.85}        & \textbf{20.14}        & -            & \textbf{49.42}        & -            & -            & \textbf{62.69}          & -              & -              & -              & -                & \textbf{38.45}                          & 20.54                     \\ \midrule
      CWSAM                            & \multirow{3}{*}{D-E-L}                                   & -          & 37.56      & \textbf{8.13}       &\textbf{6.5}        & -            & -            & -            & 37.41        & 38.59        & \textbf{8.41}         & -              & -              & -              & 36.34          & -                & 24.71                          & -                         \\
      SAM-LoRA                        &                                    & -          & \underline{49.52}      & 3.81       & \underline{4.53}       & -            & -            & -            & \underline{51.05}        & \underline{51.47}        & 4.29         & -              & -              & -              & \underline{53.08}          & -                & \underline{31.11}                          & 6.40                      \\
\rowcolor{gray!10}      MLE-SAM                             &                                    & -          & \textbf{56.02}      & \underline{4.07}       & 2.13       & -            & -            & -            & \textbf{56.45}        & \textbf{56.78}        & \underline{4.75}         & -              & -              & -              &\textbf{57.96}          & -                & \textbf{34.02}                          & 9.31                      \\ \midrule
      CMNeXt                           & \multirow{4}{*}{R-D-E-L}           & 0.86       & 0.49       & 0.66       & 0.37       & 47.06        & 9.97         & 13.75        & 2.63         & 1.73         & 2.85         & 59.03          & 59.18          & 14.73          & 39.07          & 59.18            & 20.77                          & -                         \\
      CWSAM                            &                                    & 12.3       & 35.42      & \textbf{8.16}       & \textbf{6.2}        & 23.51        & \underline{15.91}        & 15.59        & 39.2         & 37.21        & \textbf{9.11}         & 28.7           & 28.84          & \textbf{21.84}          & 44.15          & 55.43            & 25.44                          & 4.67  \\
      SAM-LoRA                        &                                    & \textbf{17.62}      & \underline{48.58}      &\underline{ 2.92}       & \underline{3.16}       & \underline{59.54}        & \textbf{17.62}        & \textbf{17.62}        & \underline{48.58}        & \underline{48.58}        & \underline{2.92}         & \underline{59.54}          & \underline{59.54}          & \underline{17.62}          & \underline{48.58}          & \underline{59.54}            & \underline{34.13}                          & 13.36 \\
\rowcolor{gray!10}      MLE-SAM                             &                                    & \underline{15.8}       & \textbf{50.28}      & 0.74       & 2.07       & \textbf{63.47}        & 15.57        & \underline{15.91}        & \textbf{50.42}        & \textbf{50.6}         & 0.86         & \textbf{63.11 }         & \textbf{64.26}          & 15.64          & \textbf{50.68}          & \textbf{64.08}            & \textbf{34.90}                          & 14.13 \\ \midrule
      \end{tabular}
\end{adjustbox}
  \end{table*}
\begin{figure*}[htp]
  \begin{center}
     \includegraphics[width=\linewidth]{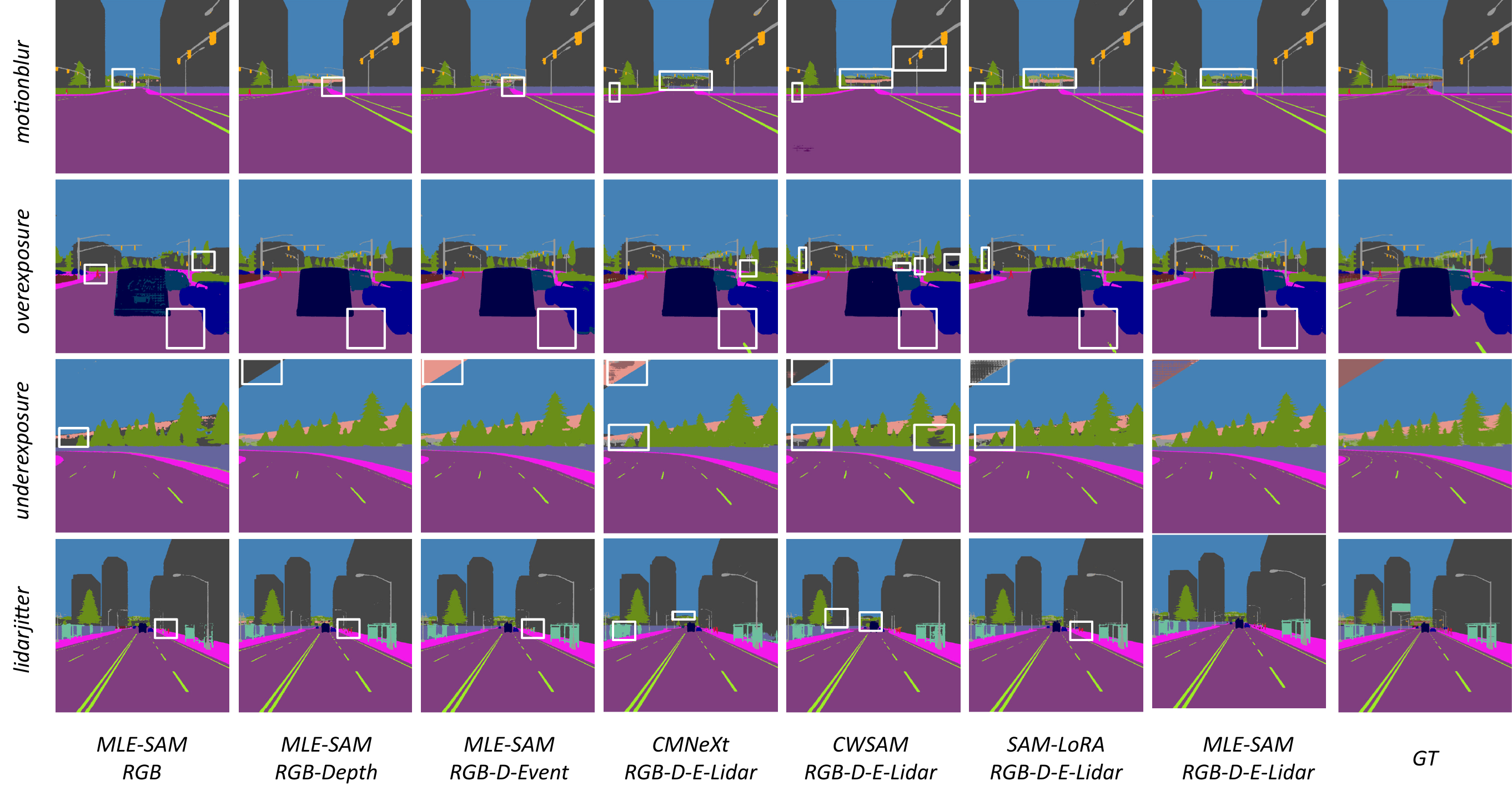}
     \caption{Comparison of semantic segmentation results on the DELIVER dataset using different methods and modalities}\label{fig:semantic}
  \end{center}
\end{figure*}
Figure~\ref{fig:semantic} presents the semantic segmentation results on the DELIVER dataset, illustrating the performance differences among various methods and modality combinations. The results indicate that integrating the R-D-E-L modality combination significantly improves segmentation accuracy and completeness compared to single-modal approaches. For example, MLE-SAM with only the RGB modality struggles to detect pedestrians under challenging conditions such as overexposure and LiDAR jitter. In contrast, the R-D-E-L combination accurately segments small objects like pedestrians. However, CWSAM and SAM-LoRA with the R-D-E-L combination exhibit suboptimal performance, particularly in segmenting buildings under overexposure, and all three methods encounter difficulties in identifying small objects during motion blur scenarios. Furthermore, CMNeXt with R-D-E-L fails to capture critical details, such as bus stations and lights, under LiDAR jitter conditions. These results underscore the robustness of MLE-SAM in leveraging comprehensive multi-modal data to achieve consistent and superior segmentation accuracy overall segmentation performance under sensor failure cases.

\subsection{Generalization Evaluation with Partial Modality Testing}
Table~\ref{tab:delivermissing} presents a comprehensive evaluation of four semantic segmentation models—CMNeXt, CWSAM, SAM-LoRA, and MLE-SAM—trained on three modality combinations: R-D-E, D-E-L, and R-D-E-L. The models were tested using the DELIVER dataset under various modality scenarios. A key limitation of CMNeXt is its dependency on the RGB modality during training, restricting its flexibility compared to CWSAM, SAM-LoRA, and MLE-SAM, which support training without RGB. Among the evaluated models, MLE-SAM consistently achieves superior performance across all training configurations. Specifically, under the R-D-E training setup, MLE-SAM achieves a mean mIoU of 38.45\%, outperforming SAM-LoRA and CWSAM by 1.58\% and 10.2\%, respectively. For the D-E-L configuration, MLE-SAM achieves 34.02\%, surpassing SAM-LoRA by 2.91\% and CWSAM by 9.31\%. Similarly, under the R-D-E-L configuration, MLE-SAM achieves the highest mean mIoU of 34.90\%, exceeding SAM-LoRA by 0.77\% and CWSAM by 9.46\%. These results highlight MLE-SAM’s effectiveness and adaptability across diverse training setups.

The impact of missing modalities during testing reveals critical insights into the interaction between dense and sparse modalities. When trained on R-D-E and tested on individual modalities, MLE-SAM demonstrates significant variability in performance, achieving 20.77\% for RGB-only testing, 48.59\% for Depth, and 4.68\% for Event. This highlights the stabilizing role of dense data, such as RGB and Depth, compared to the sparse Event modality. A similar pattern emerges under the D-E-L training setup, where Depth testing yields 56.02\%, substantially outperforming Event and LiDAR, which achieve 4.07\% and 2.13\%, respectively. For the R-D-E-L configuration, MLE-SAM demonstrates robust performance in dense testing scenarios, such as 50.28\% for Depth and 63.47\% for RGB-Depth. However, sparse-only cases, such as Event and LiDAR, result in significantly lower scores of 0.74\% and 2.07\%, respectively. These findings highlight the robustness of dense modalities in enhancing semantic segmentation performance. In contrast, while offering complementary information, sparse modalities exhibit limited effectiveness when utilized independently.

These performance patterns can be attributed to the intrinsic characteristics of dense and sparse modalities and their integration during training. Dense modalities like RGB and Depth offer rich spatial and structural information, enabling the model to learn stable and generalized features. In contrast, sparse modalities such as Event and LiDAR capture irregular and limited data, which, while applicable in specific contexts, are less reliable as standalone inputs. Training on R-D-E-L incorporates redundancy and the richness of dense data, leading to robust performance on dense subsets during testing. Conversely, reliance on sparse data during testing introduces noise, reducing predictive accuracy. Notably, excluding sparse modalities during training can mitigate these effects, as evidenced by the superior performance of RGB-Depth testing that achieves 63.47\% under the R-D-E-L training setup. This suggests that while sparse modalities provide useful complementary features, overemphasis during training can hinder model generalization. MLE-SAM’s adaptive fusion mechanism effectively integrates dense and sparse modalities, ensuring superior performance across multi-modal setups.

\begin{table*}[t!]
  \centering
  \caption{Experimental results on different modality combinations and tested under various individual and combined modality scenarios using MUSES. The modalities include Frame-camera (F), LiDAR (L), Event-camera (E)}\label{tab:musesmissing}
\renewcommand{\tabcolsep}{14pt}
  \begin{adjustbox}{width=\textwidth}
    \begin{tabular}{c|c|ccccccc|c|c}
      \midrule
      \multirow{2}{*}{\textbf{Method}} & \multirow{2}{*}{\textbf{Training}} & \multicolumn{6}{c}{\textbf{MUSES dataset}}                                     & \textbf{}    & \multirow{2}{*}{\textbf{Mean}} & \multicolumn{1}{l}{\multirow{2}{*}{$\boldsymbol{\triangle\uparrow}$}} \\ \cmidrule{3-9}
                                       &                                    & \textbf{F} & \textbf{L} & \textbf{E} & \textbf{FL} & \textbf{FE} & \textbf{LE} & \textbf{FLE} &                                & \multicolumn{1}{l}{}                  \\ \midrule
      CMNeXt                           & \multirow{4}{*}{F-L}               & 3.34       & 2.48       & -          & 47.03       & -           & -           & -            & 17.62                          & -                                     \\
      CWSAM                            &                                    & 11.61      & 2.45       & -          & 40.69       & -           & -           & -            & 18.25                          & 0.63                                  \\
      SAM-LoRA                        &                                    & \underline{53.69}      & \underline{11.79}      & -          & \underline{70.34}       & -           & -           & -            & \underline{45.27}                          & 27.66                                 \\
\rowcolor{gray!10}      MLE-SAM                             &                                    & \textbf{70.9}       & \textbf{12.96}      & -          & \textbf{75.42}       & -           & -           & -            & \textbf{53.09}                          & 35.48                                 \\ \midrule
      CMNeXt                           & \multirow{4}{*}{F-E}               & 2.72       & -          & \textbf{2.38}       & -           & 43.39       & -           & -            & 16.16                          & -                                     \\
      CWSAM                            &                                    & 25.14      & -          & 1.85       & -           & 41.77       & -           & -            & 22.92                          & 6.76                                  \\
      SAM-LoRA                        &                                    & \underline{67.96}      & -          &            & -           & \underline{67.96}       & -           & -            & \underline{45.31}                          & 29.14                                 \\
\rowcolor{gray!10}      MLE-SAM                             &                                    & \textbf{74.62}      & -          & 1.34       & -           & \textbf{74.73}       & -           & -            & \textbf{50.23}                          & 34.07                                 \\ \midrule
      CMNeXt                           & \multirow{4}{*}{F-L-E}             & 3.5        & 2.64       & \underline{2.77}       & 10.28       & 6.63        & 3.14        & 46.66        & 10.80                          & -                                     \\
      CWSAM                            &                                    & 6.48       & 4.97       & 1.98       & 13.94       & 11.59       & 2.15        & 49.98        & 13.01                          & 2.21                                  \\
      SAM-LoRA                        &                                    & \underline{48.54}      & \textbf{12.05}      & \textbf{4.37}       & \underline{70.08}       & \underline{48.54}       & \textbf{12.05}       & \underline{70.08}        & \underline{37.96}                          & 27.16                                 \\
\rowcolor{gray!10}      MLE-SAM                             &                                    & \textbf{69.67}      & 5.55       & 1.5        &\textbf{74.11}      & \textbf{69.5}        & \underline{5.55}        & \textbf{74.80}        & \textbf{42.95}                          & 32.15                                 \\ \midrule
      \end{tabular}
\end{adjustbox}
  \end{table*}


Table~\ref{tab:musesmissing} compares the performance of four models trained and tested on various modality combinations from the MUSES dataset. MLE-SAM consistently outperforms its counterparts, demonstrating robustness across modality combinations. For instance, when trained on Frame-camera and LiDAR, MLE-SAM achieves 53.09\%, surpassing SAM-LoRA by 7.82\%, CWSAM by 34.84\%, and CMNeXt by 35.47\%. This trend holds under the F-E and F-L-E scenarios, with improvements of 4.92\% and 4.99\% over SAM-LoRA, respectively.

However, missing modalities during testing significantly affect performance. For example, when trained on F-L-E but tested on sparse modalities like Event-camera or LiDAR, MLE-SAM’s scores drop to 1.5\% and 5.55\%, respectively. In contrast, when tested on dense Frame-camera data, MLE-SAM achieves 69.67\%. These results highlight the critical role of dense data in maintaining segmentation quality, as dense modalities like Frame-camera provide essential spatial continuity and detail, while sparse modalities like Event-camera and LiDAR lack this richness.
These findings reinforce the advantages of MLE-SAM’s adaptive fusion mechanism. This mechanism effectively combines multi-modal inputs to mitigate the limitations of sparse data, making it particularly suited for real-world scenarios with intermittent modality availability.

\subsection{Robustness Evaluation Under Noisy Testing Conditions}
Table \ref{tab:sam-noise-performance} evaluates the performance of three adapted SAM models, namely CWSAM, SAM-LoRA, and MLE-SAM, under Gaussian and random noise applied to four modalities. The results highlight key observations regarding the differential impact of noise on dense and sparse modalities, as well as the robustness of MLE-SAM compared to the other two models.

\begin{table}[t!]
  \centering
  \caption{Performance of Adapted SAM models under different noise types (Gaussian and Random) applied to different modalities, evaluated using mIoU.}
  \label{tab:sam-noise-performance}
  \begin{tabular}{ccccc}
    \midrule
    \textbf{Model}         & \textbf{Noise Type}       & \textbf{Modality} & \textbf{mIoU} &   $\boldsymbol{\triangle\uparrow}$     \\ \midrule
    \multirow{8}{*}{CWSAM} & \multirow{4}{*}{Gaussian} & RGB               & 29.60         & -      \\
                           &                           & Depth             & \textbf{53.87}         & -      \\
                           &                           & Event             & 54.89         & -      \\
                           &                           & LiDAR             & 54.79         & -      \\ \cmidrule{2-5} 
                           & \multirow{4}{*}{Random}   & RGB               & 23.93         & -      \\
                           &                           & Depth             & \textbf{53.18}         & -      \\
                           &                           & Event             & 54.76         & -      \\
                           &                           & LiDAR             & 54.62         & -      \\ \midrule
    \multirow{8}{*}{SAM-LoRA}   & \multirow{4}{*}{Gaussian} & RGB               & 53.83         & 24.23  \\
                           &                           & Depth             & 38.10         & -15.77 \\
                           &                           & Event             & 59.55         & 4.66   \\
                           &                           & LiDAR             & 59.54         & 4.75   \\ \cmidrule{2-5} 
                           & \multirow{4}{*}{Random}   & RGB               & 52.76         & 28.83  \\
                           &                           & Depth             & 33.56         & -19.62 \\
                           &                           & Event             & 59.55         & 4.79   \\
                           &                           & LiDAR             & 59.55         & 4.93   \\ \midrule
    \multirow{8}{*}{MLE-SAM}  & \multirow{4}{*}{Gaussian} & RGB               & \textbf{57.00}         & 27.4   \\
                           &                           & Depth             & 42.64         & -11.23 \\
                           &                           & Event             & \textbf{63.90}         & 9.01   \\
                           &                           & LiDAR             & \textbf{63.87}         & 9.08   \\ \cmidrule{2-5} 
                           & \multirow{4}{*}{Random}   & RGB               & \textbf{56.35}         & 32.42  \\
                           &                           & Depth             & 38.58         & -14.6  \\
                           &                           & Event             & \textbf{63.89}         & 9.13   \\
                           &                           & LiDAR             & \textbf{63.89}         & 9.27   \\ \midrule
    \end{tabular}
\end{table}

The analysis shows that Gaussian noise affects dense modalities (RGB, Depth) more than sparse ones (Event, LiDAR). For instance, CWSAM's RGB mIoU dropped to 29.60\% under Gaussian noise, while Depth achieved 53.87\%. Sparse modalities were less affected, with Event and LiDAR maintaining mIoU values of 54.89\% and 54.79\%. Under random noise, RGB for CWSAM dropped further to 23.93\%, and Depth to 53.18\%, while Event and LiDAR remained robust, with mIoU values of 54.76\% and 54.62\%, respectively. This highlights the resilience of sparse modalities to pixel perturbations due to their localized data nature.

MLE-SAM showed superior robustness across all modalities, outperforming CWSAM and SAM-LoRA. Under Gaussian noise, MLE-SAM's RGB mIoU was 57.00\%, significantly higher than 29.60\% for CWSAM and 53.83\% for SAM-LoRA. Sparse modalities also benefited, with Event and LiDAR achieving 63.90\% and 63.87\%, reflecting improvements of 9.01\% and 9.08\% over CWSAM, and 4.35\% and 4.33\% over SAM-LoRA. Under random noise, MLE-SAM's RGB mIoU declined slightly to 56.35\%, still outperforming CWSAM and SAM-LoRA. Event and LiDAR maintained robust mIoU values of 63.89\%, surpassing CWSAM by 9.13\% and 9.27\%, and SAM-LoRA by 4.34\% across both noise types.
Comparing Gaussian and random noise, random noise introduced higher variability for dense modalities, reducing RGB mIoU in CWSAM from 29.60\% to 23.93\%. Sparse modalities were minimally affected, with stable mIoU values across models and noise types, underscoring their robustness to global perturbations.

Overall, these results emphasize the need for modality-specific strategies for noise resilience. Dense modalities require denoising techniques, while sparse ones are naturally robust. Among the models, MLE-SAM consistently outperforms CWSAM and SAM-LoRA, validating its effectiveness for multi-modal semantic segmentation in noisy environments.

\section{Conclusion and Future Work}\label{sec:conclusion}
This paper presented MLE-SAM, a novel adaptation of the SAM2 architecture tailored for multi-modal semantic segmentation. MLE-SAM incorporates LoRA-based adaptation, a selective feature weighting mechanism, and a dual-pathway mask prediction strategy. By effectively fusing dense and sparse modalities, MLE-SAM harnesses their complementary strengths to achieve precise segmentation while maintaining robustness across diverse conditions and datasets.

Extensive experiments demonstrate that MLE-SAM consistently outperforms state-of-the-art models in terms of mIoU across various datasets and modality combinations. Notably, the model exhibits resilience in challenging scenarios, including noisy inputs and missing modalities, underscoring the advantages of its multi-modal fusion approach. Dense modalities contribute detailed spatial information crucial for high-resolution segmentation, while sparse modalities enhance robustness in adverse or resource-constrained environments.

Future research can prioritize refining the multi-modal integration through advanced pretraining techniques, noise-tolerant module designs, and adaptive attention mechanisms for sparse feature enhancement. Developing dynamic fusion strategies to balance dense and sparse modalities seamlessly can improve MLE-SAM's adaptability and effectiveness in real-world applications.






 
\clearpage
\bibliographystyle{IEEEtran}
\bibliography{ref}

\clearpage
\appendix[Implementation Details]\label{appendix:a}
The input size for all images from the three datasets is standardized to 1024$\times$1024 pixels. Image preprocessing includes data augmentation techniques such as random color jittering, horizontal flipping, Gaussian blurring, and random cropping to the target resolution of 1024$\times$1024. Following these augmentations, the images are normalized using channel-wise mean and standard deviation values.
  
The source codes of CMNeXt~\cite{zhang2023delivering} and CWSAM~\cite{pu2024classwise} were adapted for compatibility with the three datasets employed in this study. CMNeXt employs a self-query hub that dynamically selects informative features from auxiliary modalities, which are then fused with the RGB-based primary branch. Additionally, the parallel pooling mixer effectively extracts discriminative cross-modal cues. In this framework, CMNeXt relies on the RGB modality for multi-modal semantic segmentation. CWSAM introduces lightweight adapters within the SAM Vision Transformer image encoder and a novel class-wise mask decoder that generates multi-class, pixel-level predictions, tailored for semantic segmentation tasks. Furthermore, we developed SAM-LoRA, an extension of the SAM model incorporating distinct LoRA modules for each modality. Similar to MLE-SAM, we modify the SAM model by applying LoRA to the image encoder while freezing the remaining components of the SAM architecture. The LoRA adaptation is implemented by altering the query and value projections within the transformer’s attention mechanism. Specifically, the original qkv projection layer is replaced with a custom LoRA layer, which is individually trained for each modality. The masks generated by the modality-specific models are subsequently averaged to form a unified feature representation.

\begin{table}[htp]
  \centering
  \caption{Training Parameters and Configurations for MLE-SAM}\label{tab:experiment}
  \begin{tabular}{@{}ccc@{}}
  \toprule
  Parameter                      & Dataset & Value                                                                        \\ \midrule
  Image Size                     & all     & {[}1024,1024{]}                                                              \\
  Batch Size                     & all     & 6                                                                            \\
  Training Epochs                & all     & 100                                                                          \\
  Loss Function                  & all     & OhemCrossEntropy                                                             \\
  Optimizer                      & all     & AdamW                                                                        \\
  \multirow{3}{*}{Learning Rate} & DELIVER & $3\times 10^{-4}$                                                                         \\
                                 & MUSES   & $6\times 10^{-4}$                                                                         \\
                                 & MCubeS  & $8\times 10^{-3}$                                                                         \\
  Weight Decay                   & all     & 0.01                                                                         \\
  Scheduler                      & all     & \begin{tabular}[c]{@{}c@{}}Warmup Polynomial \\ Decay Scheduler\end{tabular} \\
  Scheduler Power                & all     & 0.9                                                                          \\
  Warmup Epochs                  & all     & 10                                                                           \\
  Warmup Ratio                   & all     & 0.1                                                                          \\ 
  LoRA Rank                   & all     & 32                                                                          \\ \bottomrule
  \end{tabular}
  \end{table}

Model training was conducted on an NVIDIA A100 GPU with a batch size of 6. As shown in Table~\ref{tab:experiment}, the training process employed the AdamW optimizer~\cite{Ilya2019adamw}, configured with an initial learning rate and a weight decay of 0.01, over 100 epochs. The Online Hard Example Mining Cross-Entropy loss function was used without class-specific weighting to handle imbalanced segmentation classes. To optimize learning, a warm-up polynomial learning rate scheduler was applied, with a power of 0.9. The learning rate was gradually increased during the first 10 epochs using a warm-up ratio of 0.1. The ranks of the LoRA modules were set to 32 to balance model capacity and computational efficiency.
  


 




\vfill

\end{document}